\documentclass{article}
\usepackage{ijcai26}

\usepackage{times}
\usepackage{soul}
\usepackage{url}
\usepackage[hidelinks]{hyperref}
\usepackage[utf8]{inputenc}
\usepackage[small]{caption}
\usepackage{graphicx}
\usepackage{amsmath}
\usepackage{amsthm}
\usepackage{booktabs}
\usepackage{algorithm}
\usepackage{algorithmic}
\usepackage[switch]{lineno}
\usepackage[dvipsnames]{xcolor}
\usepackage{tikz}
\usetikzlibrary{positioning}
\definecolor{promblue}{RGB}{19,61,111}
\definecolor{promteal}{RGB}{19,174,157}
\definecolor{promgreen}{RGB}{0,160,85}

\title{Benchmarks Are Not Validation: A System-Level View of Financial LLM Applications}

\author{
Burak Payzun
\and
İrem Demirtaş\and
Simona Scala\and
İrem Demirtaş\And
Elena Ferretti\\
\affiliations
Prometeia S.p.A.\\
\emails
\{burak.payzun, irem.demirtas, simona.scala, elena.ferretti, secil.arslan\}@prometeia.com
}

\begin{document}

\maketitle

\begin{abstract}
    Large language models are increasingly deployed in financial applications that combine retrieval, proprietary data, tool use, orchestration logic, monitoring, and human escalation. Yet evaluation often remains model-centric: benchmark scores, task accuracy, or one-off qualitative reviews are treated as evidence of readiness. In financial settings, this is insufficient. We take the position that financial LLM systems should not be approved for production based on benchmark performance alone. They require system-level validation evidence across the application stack: data, model design, retrieval and generation performance, agent behavior, governance, and implementation. Drawing on industry experience validating GenAI applications in financial institutions, we outline a multi-layer validation view and explain why hybrid evaluation is necessary. We discuss where LLM-as-a-judge methods are useful and why they require controls such as multiple judges, rubrics, agreement, and auditability checks. We also highlight failure modes poorly captured by static benchmarks, including retrieval failures, unfaithful generation, tool misuse, escalation errors, and operational instability. Our position is that financial LLM validation should be an ongoing system discipline rather than a one-time model scoring exercise. Validation should produce decision-ready evidence, not only scores. We conclude with a research agenda for system-aware benchmarks, agent trace validation, judge alignment protocols, and lifecycle validation standards.
\end{abstract}

\section{Introduction}

Financial LLM evaluation has improved quickly. Benchmarks such as FinBen cover financial information extraction, textual analysis, question answering, generation, risk management, forecasting, and decision-making \cite{xie2024finben}. Earlier datasets such as FinQA and ConvFinQA show that financial question answering often requires numerical and multi-step conversational reasoning over financial documents \cite{chen2021finqa,chen2022convfinqa}. These benchmarks make model comparison more systematic and expose weaknesses missed by generic evaluations, but they do not solve the validation problem faced by financial institutions.

Financial institutions are moving from experimentation to real workflows. LLM systems now summarize documents, answer customer or analyst questions, extract information for credit and lending processes, support compliance reviews, and assist internal decisions. Many are no longer simple prompt-response systems; they combine retrieval-augmented generation, proprietary knowledge bases, prompt orchestration, external tools, APIs, user interfaces, monitoring layers, and sometimes multi-agent workflows.

This shift changes what validation must mean. A benchmark score says little about whether a deployed system retrieves the right documents, preserves factual grounding, respects guardrails, handles sensitive data, escalates uncertain cases, resists prompt injection, calls tools safely, or remains stable after third-party model or API changes. In finance, these are not secondary concerns: they determine whether a system is reliable, auditable, and fit for purpose.

Banking regulation further raises the validation bar. Validation and audit functions are increasingly expected to assess not only traditional models but also LLM-based applications and AI-driven workflows. Regulatory frameworks such as Model Risk Management (MRM) guidelines and the EU AI Act reinforce the need for governance, traceability, explainability, and continuous oversight across the lifecycle.

Existing evaluation methods cover only part of this problem. Classical NLP metrics are useful for constrained tasks, but often depend on reference answers and surface similarity. Human evaluation is valuable but costly to scale. LLM-as-a-judge methods are flexible and fast, and can approximate human preferences in some open-ended settings \cite{zheng2023judging,liu2023geval}; however, they introduce prompt sensitivity, judge bias, reproducibility issues, and overconfidence. No single method is sufficient.

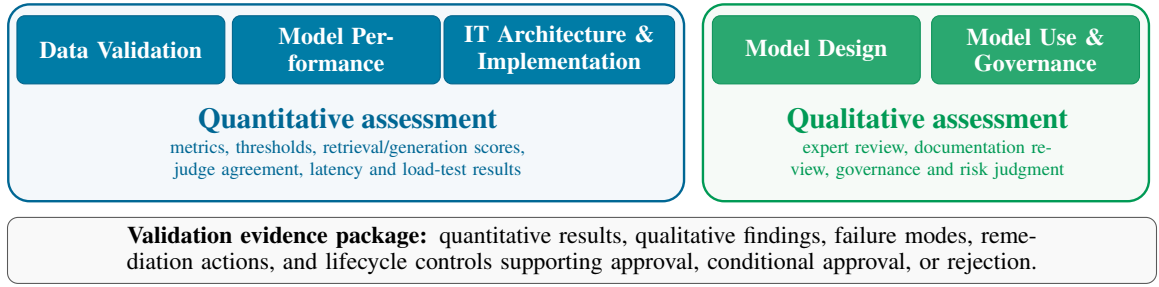
\begin{figure*}[ht]
\centering
\begin{tikzpicture}[
    quantpanel/.style={draw=MidnightBlue, fill=MidnightBlue!3, rounded corners=8pt, line width=0.9pt},
    qualpanel/.style={draw=ForestGreen, fill=ForestGreen!3, rounded corners=8pt, line width=0.9pt},
    quantpillar/.style={draw=MidnightBlue!90!black, fill=MidnightBlue!80, text=white, rounded corners=3pt, align=center, minimum width=2.75cm, minimum height=0.9cm, text width=2.50cm, font=\footnotesize\bfseries},
    qualpillar/.style={draw=ForestGreen!90!black, fill=ForestGreen!80, text=white, rounded corners=3pt, align=center, minimum width=2.75cm, minimum height=0.9cm, text width=2.50cm, font=\footnotesize\bfseries},
    evidence/.style={draw=black!65, fill=gray!5, rounded corners=4pt, align=center, minimum width=15.2cm, minimum height=0.62cm, text width=14.1cm, font=\footnotesize}
]

\draw[quantpanel] (-0.24,-1.55) rectangle (8.70,1.05);
\draw[qualpanel] (8.95,-1.55) rectangle (14.85,1.05);

\node[quantpillar] at (1.25,0.45) {Data Validation};
\node[quantpillar] at (4.10,0.45) {Model Performance};
\node[quantpillar, minimum width=3.05cm, text width=2.80cm] at (7.05,0.45) {IT Architecture \&\\Implementation};

\node[qualpillar] at (10.45,0.45) {Model Design};
\node[qualpillar] at (13.35,0.45) {Model Use \& Governance};

\node[font=\large\bfseries, text=MidnightBlue] at (4.25,-0.48) {Quantitative assessment};
\node[font=\scriptsize, align=center, text width=7.6cm, text=MidnightBlue] at (4.25,-1.00) {metrics, thresholds, retrieval/generation scores, judge agreement, latency and load-test results};

\node[font=\large\bfseries, text=ForestGreen] at (11.92,-0.48) {Qualitative assessment};
\node[font=\scriptsize, align=center, text width=5.1cm, text=ForestGreen] at (11.92,-1.00) {expert review, documentation review, governance and risk judgment};

\node[evidence] at (7.35,-2.20) {\textbf{Validation evidence package:} quantitative results, qualitative findings, failure modes, remediation actions, and lifecycle controls supporting approval, conditional approval, or rejection.};

\end{tikzpicture}
\caption{Five-pillar view of system-level validation for financial LLM applications. Data, performance, and IT architecture and implementation are primarily quantitative assessment pillars; model design and model use and governance are primarily qualitative assessment pillars. The pillars are independent assessment dimensions rather than a pipeline.}
\label{fig:five_pillars}
\end{figure*}
Institutions therefore need system-specific test sets, scenarios, and acceptance criteria reflecting their documents, workflows, users, risks, and regulatory constraints. This is difficult in practice: historical data may be limited, acceptable behavior may be unclear, annotation is costly, and synthetic cases can introduce bias. The gap between benchmark performance and evidence of system readiness remains especially large for new systems with few production traces or known failures.

We take the position that financial LLM systems should not be approved for production based on benchmark performance alone. They require system-level validation evidence across the full application stack: data, model design, retrieval and generation behavior, agentic decision logic, governance processes, and IT implementation, with human oversight calibrated to use-case risk. Validation requirements are also use-case dependent. A low-risk internal assistant, a RAG-based knowledge search system, a customer-facing chatbot, and a system supporting creditworthiness assessment have different risk profiles and validation expectations.

We organize this validation view around five independent pillars: data, model design, performance, model use and governance, and IT architecture and implementation. These pillars are not a pipeline or hierarchy; they are complementary assessment dimensions. Validation evidence should also distinguish quantitative assessment, mainly supporting data, performance, and IT implementation, from qualitative assessment, mainly supporting model design and governance. Both forms of evidence feed a structured validation evidence package that supports approval, conditional approval, or rejection. Figure~\ref{fig:five_pillars} summarizes this view.

This paper makes three contributions. First, it clarifies why benchmark-centric evaluation is insufficient for deployed financial LLM systems. Second, it proposes a system-level validation view covering data, model design, performance, agent behavior, governance, and implementation. Third, it identifies research directions for financial LLM validation, including trace-level agent evaluation, auditable LLM-as-a-judge protocols, and lifecycle validation standards.

\section{Related Work}

Financial benchmark suites evaluate LLMs across broad task families such as information extraction, question answering, forecasting, risk management, and decision-making \cite{xie2024finben}, but benchmark coverage is not deployment validation.

Surveys of financial foundation models catalog open challenges, compliance, hallucination, non-stationarity, and deployment cost---across language, time-series, and visual-language models \cite{chen2025finfm}. Benchmarking DeepSeek-R1 on financial QA shows strong accuracy but leaves a deployment gap from small single-choice datasets, persistent hallucination under regulation, and missing multimodality \cite{liu2025deepseek}. Both frame these as model-level concerns; we treat them as system-level validation requirements.

RAG evaluation moves beyond final-answer scoring by separating retrieval quality, context relevance, answer relevance, and faithfulness; RAGAS, for example, provides reference-free metrics for these modular pipelines \cite{es2024ragas}. This matters in finance, where applications built on internal document collections make grounding failures a material risk.

LLM-as-a-judge methods can approximate human preferences in open-ended settings \cite{zheng2023judging} and improve correlation with human judgments on selected NLG tasks \cite{liu2023geval}, but they exhibit position, verbosity, authority, and self-preference biases (Table~\ref{tab:llm_judge_pitfalls}) \cite{zheng2023judging,chen2024humans}. Panels or juries of judges reduce single-model dependence \cite{verga2024replacing}, yet still require auditability and agreement checks.

\begin{table*}[h]
\centering
\small
\renewcommand{\arraystretch}{1.2}
\begin{tabular}{p{3.2cm} p{4.3cm} p{4.3cm} p{4.8cm}}
\hline
\textbf{Pitfall / Bias} & \textbf{Description} & \textbf{Typical Failure Mode} & \textbf{Recommended Mitigation} \\
\hline
Position bias & Order affects ratings. & First response favored. & Randomize order; evaluate both directions. \\
Verbosity bias & Length is mistaken for quality. & Verbose weak answers outrank concise accurate ones. & Include conciseness criteria; normalize length. \\
Authority bias & Confident references inflate scores. & Hallucinated citations are rewarded. & Separate factual verification from style. \\
Sentiment / compassion bias & Politeness affects scores. & Friendly responses receive inflated ratings. & Use domain rubrics focused on substance. \\
Self-enhancement bias & Judges prefer same-family outputs. & Provider favoritism distorts comparisons. & Use heterogeneous evaluator panels. \\
Agreement illusion & Judges share blind spots. & Consensus forms around wrong answers. & Measure disagreement and diversify judges. \\
Prompt sensitivity & Prompt wording changes scores. & Same response receives inconsistent ratings. & Version-control prompts; test prompt robustness. \\
Rubric ambiguity & Vague criteria enable subjective scoring. & ``Helpful'' or ``good'' is interpreted inconsistently. & Use operationalized rubrics with examples. \\
Chain-of-thought bias & Reasoning presentation affects scores. & Persuasive reasoning is rewarded when wrong. & Blind judges to hidden reasoning when appropriate. \\
Style-over-substance effect & Fluency is mistaken for correctness. & Well-written hallucinations score highly. & Combine judge scores with factuality checks. \\
Calibration instability & Scales differ across judges. & One judge's ``4'' equals another's ``2''. & Calibrate against human-rated anchors. \\
Low reproducibility & Scores vary across runs or versions. & Repeated evaluations disagree. & Fix evaluator versions and decoding parameters. \\
Distribution shift & Generic judges fail in specialized domains. & Finance or legal nuance is missed. & Use domain evaluators and SME review. \\
Evaluator hallucination & Judges fabricate rationales. & Arbitrary scores appear justified. & Treat rationales as evidence, not truth. \\
Overfitting to the judge & Systems optimize to evaluator preferences. & Models learn to game judges. & Rotate judges and add human spot checks. \\
Averaging away failures & Aggregates hide edge cases. & Critical failures disappear in averages. & Report worst cases and category results. \\
Weak adversarial robustness & Judges fail under adversarial inputs. & Unsafe outputs evade detection. & Red-team the evaluator. \\
Weak human alignment & Human correlation is incomplete. & Judges disagree with expert reviewers. & Benchmark against expert annotators. \\
Metric monoculture & LLM judges become the only metric. & Qualitative scores replace validation. & Combine LLM judges with classical, retrieval, and rule-based checks. \\
\hline
\end{tabular}
\caption{Common pitfalls and shortcomings of LLM-as-a-Judge evaluation in practice. Although LLM judges provide scalable qualitative evaluation, they remain vulnerable to systematic biases, instability, and evaluator-specific failure modes.}
\label{tab:llm_judge_pitfalls}
\end{table*}

Agent evaluation is another emerging line of work. AgentBench evaluates LLMs as agents in interactive environments and identifies long-term reasoning, decision-making, and instruction following as central obstacles \cite{liu2025agentbenchevaluatingllmsagents}. StableToolBench focuses on the difficulty of stable tool-use evaluation when LLMs interact with external tools and APIs \cite{guo2025stabletoolbenchstablelargescalebenchmarking}. AgentDiagnose argues that final-task success leaves agent decision processes opaque and proposes trajectory-level diagnosis \cite{ou-etal-2025-agentdiagnose}. This matters for finance because errors in tool selection, parameter passing, permissions, or escalation can be more important than the fluency of the final response.

\begin{table*}[h]
\centering
\small
\renewcommand{\arraystretch}{1.15}
\begin{tabular}{p{3.2cm} p{4.2cm} p{4.2cm} p{4.2cm}}
\hline
\textbf{Validation Level} & \textbf{Evidence Used} & \textbf{Typical Checks} & \textbf{Failure Diagnosed} \\
\hline
Black-box outcome & User input and final output & Task success, correctness, refusal, escalation, tone, safety & Visible user-facing failure \\
Grey-box behavior & Output plus retrieved context, confidence, tool summaries, logs & Grounding, evidence use, uncertainty handling, escalation path & Misuse of partial system evidence \\
White-box trace & Full trajectory, tool calls, parameters, observations, handoffs & Tool choice, parameter validity, permissions, retries, loop detection & Unsafe or inefficient intermediate behavior \\
Ablation / replay & Controlled model, retrieval, prompt, and tool variants & Base-vs-system attribution, regression and migration testing & Root cause across components \\
\hline
\end{tabular}
\caption{Agent validation should combine black-box, grey-box, white-box, and ablation-based evidence. Final-answer correctness is necessary but insufficient when financial LLM systems can retrieve, route, call tools, or execute workflows.}
\label{tab:agent_validation_levels}
\end{table*}

A useful distinction can be made between \textit{white-box}, \textit{black-box}, and \textit{grey-box} evaluation approaches, as also shown in Table~\ref{tab:agent_validation_levels}:
\begin{itemize}
    \item \textit{White-box evaluation}: uses traces, metadata, or internal outputs to assess intermediate behavior such as tool selection, parameter correctness, evidence use, reasoning trajectories, or task decomposition. This is reflected in diagnostic datasets and benchmarks that inspect process rather than only final answers \cite{mialon2023gaiabenchmarkgeneralai,wang2022supernaturalinstructionsgeneralizationdeclarativeinstructions,Wolfson2020Break}.

    \item \textit{Black-box evaluation}: evaluates only the user input and final output. Typical checks include outcome correctness, robustness to prompt transformations, repeated-run consistency, safety, and refusal behavior.

    \item \textit{Grey-box evaluation}: combines outcome evaluation with partial internal information, such as retrieved passages, confidence signals, tool-call summaries, or escalation logs. It is useful when assessing whether the system recognized missing, conflicting, or unsafe conditions.
\end{itemize}

This distinction is important in finance because a correct-looking final answer can conceal unsafe intermediate behavior, policy violations, or incorrect tool usage.

The gap is therefore clear: existing work provides useful components for evaluation, but financial institutions need an integrated validation view that connects benchmark performance, RAG evaluation, judge reliability, agent traces, governance, security, and production implementation.

\section{The Limits of Benchmark-Centric Evaluation}

Benchmarks are useful for comparability, reporting, and model selection, but financial LLM systems fail in ways that benchmark scores do not capture.

First, benchmarks usually evaluate models or tasks in isolation. Deployed financial applications also include ingestion, chunking, embedding, retrieval, prompt construction, generation, post-processing, logging, feedback collection, and escalation. A failure in any component can produce an incorrect or unsafe output even when the underlying model is strong.

Second, financial tasks are context-specific. Public benchmarks may not reflect an institution's documents, products, regulatory environment, language mix, risk appetite, or operational constraints. A model that performs well on general financial question answering may still fail on internal policies, local banking terminology, or specific reporting templates.

Third, even finance-specific benchmarks remain bounded by task format, source material, annotation strategy, answer type, and evaluation protocol. FinBen covers broad financial tasks, and FinanceBench provides open-book question answering over company filings \cite{xie2024finben,islam2023financebench}; however, both still convert financial work into fixed test items, reference answers, and simplified acceptance conditions. Deployment is broader: systems must operate over proprietary taxonomies, changing product definitions, multilingual documents, ambiguous requests, incomplete evidence, and downstream business processes.

Benchmarks should therefore be treated as sampling instruments, not complete validation environments. They can show performance on known task families, but not whether a specific institution's application has been validated against its operational universe, control environment, and risk appetite.

Fourth, agentic systems introduce sequential failure modes. Once an LLM can call tools, route requests, invoke APIs, decide whether to escalate, or coordinate with other agents, validation must cover traces and decision sequences. Prompt-level accuracy does not guarantee safe system behavior.

Fifth, rapid model replacement can invalidate prior validation. A newer model may improve benchmark accuracy while changing refusal behavior, citation style, tool-use reliability, latency, cost, calibration, or prompt sensitivity. In finance, migration to a new foundation model, embedding model, retriever, prompt template, tool schema, guardrail, or orchestration layer should be treated as a controlled change and trigger targeted regression testing against accepted cases, known failures, and high-risk scenarios.

RAG systems also require component-level evaluation. Retrieval quality, context relevance, and generation behavior should be assessed separately, because upstream errors propagate into final outputs.

For these reasons, benchmark-centric evaluation should be an input to validation, not validation itself.

\section{A System-Level View of Financial LLM Validation}

The following subsections expand the five pillars introduced in Figure~\ref{fig:five_pillars}. The exact implementation may vary by institution and use case, but the validation logic should cover the full application lifecycle and produce a structured evidence package rather than a single score.

\subsection{Data Validation}

Data validation assesses the quality, relevance, safety, and structure of training data where applicable, test data, evaluation datasets, and indexed knowledge bases used in RAG systems. It should cover corpus-level statistics, language scope, duplication, cleaning, toxicity or unsafe-content screening, privacy checks, and representativeness for the intended use case. In RAG systems, the indexed corpus is part of the model's effective knowledge environment; poor data health can cause poor retrieval, hallucinated answers, or unsafe outputs even when the generator is capable.

Data augmentation can support validation through perturbations, paraphrases, entity replacements, and synthetic edge cases that test robustness, fairness, and wording sensitivity. It must also be controlled: generated or transformed data can introduce artifacts, distort meaning, or create misleading evaluation results if not documented and checked.

\subsection{Model Design Validation}

Model design validation asks whether the architecture and development strategy are fit for the task. It covers the foundation model, embedding model, retrieval design, prompt strategy, fine-tuning approach, model-size reduction, and human feedback or escalation mechanisms.

The key question is not whether the system uses the most advanced model, but whether the design is fit for purpose. A smaller model may suit a constrained classification task; RAG may be appropriate when answers must be grounded in internal documents; fine-tuning may be justified for domain adaptation; and prompt engineering may be sufficient for simpler workflows. Validation should also check whether assumptions on task scope, user behavior, source data, escalation rules, model limitations, regulatory constraints, and third-party dependencies are documented.

\subsection{Model Performance Validation}

Performance validation should begin with the validation objective, not a catalogue of metrics. Institutions should first define what correct, acceptable, incomplete, unsafe, and escalation-worthy behavior means for each use case, considering the business task, users, error consequences, ground-truth availability, explainability needs, and permitted autonomy. Only then should evaluators select quantitative, qualitative, human, automated, or hybrid evidence.

A comprehensive evaluation should cover factual correctness, grounding, completeness, consistency, uncertainty handling, refusal behavior, robustness to paraphrase or noisy input, prompt sensitivity, privacy and confidentiality, fairness and bias where relevant, operational efficiency, and downstream business impact. This is consistent with holistic evaluation, which calls for assessment across multiple scenarios and desiderata rather than isolated accuracy measures \cite{liang2023helm}, and with surveys emphasizing reliable, reproducible, and robust evaluation before deployment \cite{laskar2024systematic}.

Validation evidence should combine complementary methods. Structured tasks may use deterministic rules; RAG systems require evidence-level checks separating retrieval from answer quality; open-ended generation may require expert review, rubrics, or controlled LLM-as-a-judge evaluation. High-risk use cases should include adversarial, edge-case, and regression tests, while production systems require monitoring because offline results may not hold under changing users, documents, prompts, providers, or market conditions.

Performance should not be reported only through aggregate scores, since averages can hide severe rare failures. Reports should include category-level results, worst-case failures, error taxonomies, confidence or escalation analysis, and remediation evidence. In operational settings, results are often summarized as acceptable, requires remediation, or not acceptable. The central question is whether the system provides sufficient evidence of fitness for the intended financial use case.

\subsection{LLM-as-a-Judge: Necessary but Not Sufficient}

LLM-as-a-judge methods are increasingly used because they scale qualitative assessment where classical metrics are insufficient. This is relevant for financial LLMs, whose outputs are often open-ended, context-dependent, and difficult to compare with a single reference answer. A response may be correct but incomplete, fluent but insufficiently grounded, concise but missing risk information, or helpful but inappropriate under policy constraints. LLM judges can therefore provide rubric-based signals for correctness, completeness, relevance, conciseness, language quality, tone, bias, and safety. MT-Bench, Chatbot Arena, and G-Eval show that strong LLM evaluators can align with human preferences or judgments in some settings \cite{zheng2023judging,liu2023geval}.

In financial validation, LLM-as-a-judge should be a structured evaluator, not a source of truth. It can assess answer quality, RAG faithfulness, instruction and communication alignment, and safety or policy issues such as privacy violations, harmfulness, toxicity, defamation, inappropriate specialized advice, intellectual property concerns, unsafe tool use, and misuse. These dimensions are especially relevant in lending, compliance, AML, fraud, credit risk, and customer-facing workflows.

Judge scores should be diagnostic evidence rather than aggregate benchmark results. Their value is granularity: a system may perform well on tone and fluency while failing on factual grounding, privacy handling, escalation, or policy compliance. In practice, judge-based evaluation often uses structured rubrics, such as 1--5 scales, with explicit thresholds for dimensions such as correctness, safety, and bias, calibrated against human-reviewed examples.

LLM judges should not be treated as oracles. They can be sensitive to prompt wording, rubrics, response order, verbosity, confidence, sentiment, authority signals, and model family. Prior work documents position, verbosity, and self-enhancement biases \cite{zheng2023judging}, while later studies show that both human and LLM judges are vulnerable to judgment biases and perturbations \cite{chen2024humans}. Judges may reward plausible but wrong explanations or agree for the wrong reasons when they share training distributions or model-family biases.

A responsible setup should include explicit rubrics, structured outputs, documented prompts, fixed evaluator versions, controlled decoding parameters, multiple and diverse judges where possible, and agreement analysis. Evaluator juries are consistent with recent work on panels of diverse models \cite{verga2024replacing}. Ordinal scores may require weighted agreement, ranked outputs may require rank correlation, and high disagreement should trigger expert review rather than be averaged away. For high-risk financial use cases, judge outputs should be calibrated against human-rated anchors and periodically checked by subject-matter experts.

The goal is to use LLM judges as scalable evaluators within a controlled validation process. In finance, they should be auditable, reproducible, subordinate to the validation objective, and complementary to deterministic checks, retrieval and faithfulness metrics, adversarial testing, expert review, and lifecycle monitoring.

\subsection{Agent and Tool Validation}

Agentic financial LLM systems require a separate validation layer because they do not only generate text. They may classify intent, retrieve evidence, select tools, call APIs, route tasks, escalate, or execute multi-step workflows. Validation should therefore begin by classifying the type and degree of agency: a low-agency RAG assistant, a tool-calling assistant, a workflow agent that updates records, and a multi-agent system have different risk profiles.

Black-box evaluation is necessary but insufficient. A final answer may look correct even if the system used the wrong source, called an unnecessary tool, passed unsafe parameters, ignored permissions, or skipped escalation. Conversely, an incorrect answer may originate from the base model, retriever, tool schema, API failure, orchestration logic, memory, or guardrails.

Agent validation should include trace- and component-level evidence. At minimum, it should examine task identification, decomposition, retrieval, tool choice, parameter validity, tool-error handling, permission boundaries, side effects, escalation of uncertain or high-risk cases, and faithfulness of the final answer. This aligns with agent benchmarks and diagnostic work on real-world assistant tasks, interactive environments, stable tool use, and trajectory-level diagnosis \cite{mialon2023gaia,liu2024agentbench,guo2024stabletoolbench,ou2025agentdiagnose}.

A practical design should combine black-box, grey-box, and white-box checks. Black-box tests assess the final outcome. Grey-box tests use partial internal information such as retrieved passages, confidence scores, tool-call summaries, or escalation logs. White-box tests inspect full traces, including observations, tool calls, parameters, retries, failures, and handoffs. The more authority the agent has to affect real systems, the more important grey-box and white-box validation become.

Ablation and replay tests help separate base-model performance from system performance. Tasks can be rerun with retrieval disabled, fixed context, mocked tool outputs, alternative model versions, or golden tool-call sequences. This attribution matters because remediation differs: model weaknesses may require prompt redesign or model replacement, while tool-use failures may require schema constraints, permissions, fallback logic, or human approval.

Agent-level acceptance tests should cover normal workflow completion, ambiguous requests, missing or conflicting evidence, permission boundaries, unsafe-action prevention, escalation, repeated-run stability, loop detection, cost and latency limits, tool-failure recovery, and prompt-injection scenarios. High-impact workflows should also use canary releases, shadow mode, and human-in-the-loop review before full deployment.

\subsection{Model Use, Governance, and Lifecycle Validation}

\subsubsection{Regulatory Alignment}
Regulatory compliance should not be treated as an external checklist applied after technical evaluation. For financial LLM systems, regulation is part of the validation objective because the relevant object of assessment is the full application stack: data, model design, retrieval and generation behavior, agent logic, governance, and implementation controls.

The EU Artificial Intelligence Act provides a key regulatory baseline for LLM applications deployed in the European financial sector. Its risk-based approach requires application-specific classification by intended purpose, value-chain role, autonomy, affected users, and potential impact. A generic internal assistant, a policy-search RAG system, an analyst-support tool, and a system contributing to creditworthiness assessment may therefore face different obligations even if they use the same foundation model. Commission guidelines on AI-system definition, prohibited practices, and GPAI obligations further clarify the scope of the Act, while the GPAI Code of Practice supports compliance for general-purpose AI model providers. At the same time, the Digital Omnibus package shows that AI Act implementation remains a moving regulatory target, with the May 2026 provisional agreement introducing simplification measures and delayed application dates for certain high-risk obligations.

For validation, these requirements translate into practical evidence rather than legal formality. The process should verify use-case classification, prohibited-use screening, provider and deployer responsibilities, documentation, auditability, traceability, logging, human oversight, escalation rules, bias and privacy controls, incident management, and revalidation triggers after material changes. Regulatory aspects should therefore be assessed within broader controls on data, methodology, process, governance, privacy, bias, human-in-the-loop design, monitoring, and model use. These regulatory requirements should be interpreted as validation evidence requirements rather than abstract principles.
\subsubsection{Lifecycle Validation}

Financial LLM systems require governance after deployment. Validation should cover versioning of models, prompts, data, retrieval indexes, rules, traces, and evaluation tests, as well as monitoring, feedback collection, change management, human-in-the-loop processes, and periodic revalidation.

Deployment controls such as A/B testing, shadowing, dark launches, and canary releases allow institutions to compare versions, detect degradation, and limit failure impact. Feedback mechanisms should capture both user satisfaction and substantive issues such as incorrect answers, missing evidence, unsafe content, or failed escalation.

Lifecycle validation is necessary because LLM systems are unstable in ways traditional software often is not. Foundation models may be updated, deprecated, or replaced; user behavior, retrieval corpora, prompts, and regulatory expectations may change. Validation must therefore be continuous.

\subsection{IT Architecture and Implementation Validation}

Implementation readiness is integral to GenAI validation. A financial LLM system is not fit for purpose solely because it produces accurate answers in controlled tests; it must also be reliable, scalable, observable, secure, and maintainable in production.

Validation should assess whether the architecture supports intended use under normal, peak, and exceptional conditions. This includes response-time analysis, stress and load testing, concurrency and burst testing, failure testing, dependency review, rate-limit assessment, fallback mechanisms, and third-party availability. For RAG and agentic systems, it should also cover retrieval and tool-call latency, number of model or tool calls per task, caching, retry logic, timeout handling, and degraded-service modes.

Operational weaknesses may create model risk even when model-level performance is acceptable. Systems can fail if peak loads are not handled, latency exceeds requirements, API rate limits are reached, third-party models become unavailable, or deprecated components are not replaced through controlled change. The technology stack should therefore be reviewed for scalability, security vulnerabilities, dependency management, version compatibility, and end-of-life risks. Updates to models, APIs, libraries, vector databases, orchestration tools, or cloud components should trigger regression testing on output quality, latency, cost, and risk controls.

Logs are a core validation artifact. They support auditability, incident investigation, monitoring, regulatory review, and improvement. Logging should capture prompts or prompt identifiers, retrieved context references, model and data versions, component-level response times, errors, fallback activation, tool calls, user feedback, and evaluation results, while avoiding unnecessary privacy, confidentiality, and security risks.

\section{Failure Modes That Require System-Level Validation}

As supported by our empirical experience, a system-level approach is necessary because many failures arise between components rather than inside a single model.

A RAG system may retrieve irrelevant but semantically similar documents, causing a confident but unsupported answer. It may retrieve the right documents but omit key conditions, or perform well on normal inputs but fail under paraphrased, noisy, multilingual, or adversarial inputs. RAG evaluation is useful because it distinguishes retrieval, context, answer relevance, and faithfulness instead of collapsing them into one final score \cite{es2024ragas}.

Security failures also cut across layers. Prompt injection can be introduced directly by a user or indirectly through retrieved documents. Data poisoning can affect retrieval and generation. Excessive agency can turn a model error into an unauthorized action. Poor output handling can pass unsafe content or malformed structured output to downstream systems.

These risks show why guardrails are part of system-level validation, not optional add-ons. Prompt injection, retrieval poisoning, unsafe tool calls, policy evasion, and escalation errors can propagate across retrieval, generation, orchestration, and implementation. In agentic settings, they may also affect tool selection, parameter passing, permission boundaries, and downstream actions. Guardrail architectures should therefore combine preventive controls, runtime monitoring, and post-generation validation.

Validation should assess not only whether guardrails exist, but whether they remain effective under realistic and adversarial conditions. Relevant controls include prompt filtering, retrieval sanitization, output validation, permission boundaries, policy enforcement, escalation rules, human approval workflows, logging, and fallback mechanisms. These failures are hard to detect through static benchmarks because they emerge from interactions among data, retrieval, generation, agent traces, controls, monitoring, and implementation.

\section{Research Agenda}

System-level validation for financial LLMs remains underdeveloped. We identify several directions for the research community.

First, financial LLM benchmarks should move beyond static question-answer pairs. Future benchmarks should include retrieval corpora, evidence requirements, tool-use traces, escalation scenarios, adversarial documents, and operational constraints.

Second, agent validation needs better trace-level methods. Finance requires domain-specific criteria for tool selection, parameter correctness, permission compliance, escalation, and failure recovery.

Third, LLM-as-a-judge protocols need standardization through reusable rubrics, judge alignment metrics, disagreement handling rules, and auditable reporting standards.

Fourth, financial-domain red-teaming should become a validation discipline covering prompt injection, retrieval poisoning, privacy leakage, misinformation, excessive agency, and unbounded consumption in realistic workflows.

Fifth, lifecycle validation should be integrated into financial LLM evaluation. Model updates, prompt changes, data drift, index refreshes, and third-party dependency changes should trigger structured revalidation.

Sixth, regulation-aware validation should clarify how evolving requirements, especially under the EU AI Act, can be operationalized in technical frameworks. This requires linking use-case risk classification, governance obligations, auditability, human oversight, and revalidation triggers to observable system behavior.

Finally, validation research should address the gap between technical metrics and institutional decision-making. Financial institutions need outputs that support approval decisions, risk classification, remediation planning, and monitoring.

\section{Conclusion}

Financial LLM systems should not be validated by benchmarks alone. This becomes critical as banks deploy RAG-based, tool-using, and agentic applications in workflows such as lending, compliance, anti-money laundering, fraud monitoring, and credit risk assessment. The relevant question is not only whether a foundation model performs well on a public benchmark, but whether the deployed system is grounded, reliable, auditable, secure, and fit for its intended financial purpose.

The regulatory nature of banking makes this shift unavoidable. Financial institutions operate under model risk management expectations, internal controls, and emerging AI regulations that require structured and evidence-based assessment. As LLM applications enter decision-support processes, validation and audit teams need protocols combining quantitative metrics, qualitative expert review, trace-level analysis, governance checks, and lifecycle monitoring. Evaluation should therefore move from isolated model scoring toward repeatable validation evidence that supports approval, risk classification, remediation, and oversight.

This paper takes the position that financial LLM validation is a system-level discipline. The object of validation is the full application stack: data, model design, retrieval and generation behavior, agent and tool use, guardrails, governance, and IT implementation. Hybrid evaluation is necessary because classical metrics, human review, and LLM-as-a-judge each provide useful but incomplete evidence. In finance, LLM-as-a-judge should be controlled, auditable, and subordinate to the validation objective.

Moving beyond benchmarks is not a rejection of benchmarks. They remain useful for model comparison and initial screening. Responsible deployment in finance, however, requires evidence about the whole system under realistic, adversarial, changing, and regulated conditions. As financial institutions adopt increasingly agentic LLM systems, validation should evolve into an ongoing discipline connecting technical performance, operational resilience, governance, auditability, and regulatory fitness across the lifecycle.

\bibliographystyle{named}
\bibliography{ijcai26}

@misc{mialon2023gaiabenchmarkgeneralai,
      title={GAIA: a benchmark for General AI Assistants}, 
      author={Grégoire Mialon and Clémentine Fourrier and Craig Swift and Thomas Wolf and Yann LeCun and Thomas Scialom},
      year={2023},
      eprint={2311.12983},
      archivePrefix={arXiv},
      primaryClass={cs.CL},
      url={https://arxiv.org/abs/2311.12983}, 
}

@misc{wang2022supernaturalinstructionsgeneralizationdeclarativeinstructions,
    title={Super-NaturalInstructions: Generalization via Declarative Instructions on 1600+ NLP Tasks}, 
    author={Yizhong Wang and Swaroop Mishra and Pegah Alipoormolabashi and Yeganeh Kordi and Amirreza Mirzaei and Anjana Arunkumar and Arjun Ashok and Arut Selvan Dhanasekaran and Atharva Naik and David Stap and Eshaan Pathak and Giannis Karamanolakis and Haizhi Gary Lai and Ishan Purohit and Ishani Mondal and Jacob Anderson and Kirby Kuznia and Krima Doshi and Maitreya Patel and Kuntal Kumar Pal and Mehrad Moradshahi and Mihir Parmar and Mirali Purohit and Neeraj Varshney and Phani Rohitha Kaza and Pulkit Verma and Ravsehaj Singh Puri and Rushang Karia and Shailaja Keyur Sampat and Savan Doshi and Siddhartha Mishra and Sujan Reddy and Sumanta Patro and Tanay Dixit and Xudong Shen and Chitta Baral and Yejin Choi and Noah A. Smith and Hannaneh Hajishirzi and Daniel Khashabi},
    year={2022},
    eprint={2204.07705},
    archivePrefix={arXiv},
    primaryClass={cs.CL},
    url={https://arxiv.org/abs/2204.07705}, 
}

@article{Wolfson2020Break,
  title={Break It Down: A Question Understanding Benchmark},
  author={Wolfson, Tomer and Geva, Mor and Gupta, Ankit and Gardner, Matt and Goldberg, Yoav and Deutch, Daniel and Berant, Jonathan},
  journal={Transactions of the Association for Computational Linguistics},
  year={2020},
}

@misc{guo2025stabletoolbenchstablelargescalebenchmarking,
      title={StableToolBench: Towards Stable Large-Scale Benchmarking on Tool Learning of Large Language Models}, 
      author={Zhicheng Guo and Sijie Cheng and Hao Wang and Shihao Liang and Yujia Qin and Peng Li and Zhiyuan Liu and Maosong Sun and Yang Liu},
      year={2025},
      eprint={2403.07714},
      archivePrefix={arXiv},
      primaryClass={cs.CL},
      url={https://arxiv.org/abs/2403.07714}, 
}

@misc{liu2025agentbenchevaluatingllmsagents,
      title={AgentBench: Evaluating LLMs as Agents}, 
      author={Xiao Liu and Hao Yu and Hanchen Zhang and Yifan Xu and Xuanyu Lei and Hanyu Lai and Yu Gu and Hangliang Ding and Kaiwen Men and Kejuan Yang and Shudan Zhang and Xiang Deng and Aohan Zeng and Zhengxiao Du and Chenhui Zhang and Sheng Shen and Tianjun Zhang and Yu Su and Huan Sun and Minlie Huang and Yuxiao Dong and Jie Tang},
      year={2025},
      eprint={2308.03688},
      archivePrefix={arXiv},
      primaryClass={cs.AI},
      url={https://arxiv.org/abs/2308.03688}, 
}

@inproceedings{ou-etal-2025-agentdiagnose,
    title = "{A}gent{D}iagnose: An Open Toolkit for Diagnosing {LLM} Agent Trajectories",
    author = "Ou, Tianyue  and
      Guo, Wanyao  and
      Gandhi, Apurva  and
      Neubig, Graham  and
      Yue, Xiang",
    editor = {Habernal, Ivan  and
      Schulam, Peter  and
      Tiedemann, J{\"o}rg},
    booktitle = "Proceedings of the 2025 Conference on Empirical Methods in Natural Language Processing: System Demonstrations",
    month = nov,
    year = "2025",
    address = "Suzhou, China",
    publisher = "Association for Computational Linguistics",
    url = "https://aclanthology.org/2025.emnlp-demos.15/",
    doi = "10.18653/v1/2025.emnlp-demos.15",
    pages = "207--215",
    ISBN = "979-8-89176-334-0"
}

@article{islam2023financebench,
  title={FinanceBench: A New Benchmark for Financial Question Answering},
  author={Islam, Pranab and Kannappan, Anand and Kiela, Douwe and Qian, Rebecca and Scherrer, Nino and Vidgen, Bertie},
  journal={arXiv preprint arXiv:2311.11944},
  year={2023}
}

@article{liang2023helm,
  title={Holistic Evaluation of Language Models},
  author={Liang, Percy and Bommasani, Rishi and Lee, Tony and Tsipras, Dimitris and Soylu, Dilara and Yasunaga, Michihiro and Zhang, Yian and others},
  journal={Transactions on Machine Learning Research},
  year={2023}
}

@inproceedings{laskar2024systematic,
  title={A Systematic Survey and Critical Review on Evaluating Large Language Models: Challenges, Limitations, and Recommendations},
  author={Laskar, Md Tahmid Rahman and Alqahtani, Sawsan and Bari, M Saiful and Rahman, Mizanur and Khan, Mohammad Abdullah Matin and Khan, Haidar and Jahan, Israt and Bhuiyan, Amran and Tan, Chee Wei and Parvez, Md Rizwan and Hoque, Enamul and Joty, Shafiq and Huang, Jimmy Xiangji},
  booktitle={Proceedings of the 2024 Conference on Empirical Methods in Natural Language Processing},
  pages={13785--13816},
  year={2024}
}

@article{mialon2023gaia,
  title={GAIA: A Benchmark for General AI Assistants},
  author={Mialon, Grégoire and Fourrier, Clémentine and Swift, Craig and Wolf, Thomas and LeCun, Yann and Scialom, Thomas},
  journal={arXiv preprint arXiv:2311.12983},
  year={2023}
}

@inproceedings{guo2024stabletoolbench,
  title={StableToolBench: Towards Stable Large-Scale Benchmarking on Tool Learning of Large Language Models},
  author={Guo, Zhicheng and Cheng, Sijie and Wang, Hao and Liang, Shihao and Qin, Yujia and Li, Peng and Liu, Zhiyuan and Sun, Maosong and Liu, Yang},
  booktitle={Findings of the Association for Computational Linguistics: ACL 2024},
  pages={11143--11156},
  year={2024}
}

@inproceedings{chen2021finqa,
  title = {{F}in{QA}: A Dataset of Numerical Reasoning over Financial Data},
  author = {Chen, Zhiyu and Chen, Wenhu and Smiley, Charese and Shah, Sameena and Borova, Iana and Langdon, Dylan and Moussa, Reema and Beane, Matt and Huang, Ting-Hao and Routledge, Bryan and Wang, William Yang},
  booktitle = {Proceedings of the 2021 Conference on Empirical Methods in Natural Language Processing},
  pages = {3697--3711},
  year = {2021},
  month = nov,
  address = {Online and Punta Cana, Dominican Republic},
  publisher = {Association for Computational Linguistics},
  doi = {10.18653/v1/2021.emnlp-main.300},
  url = {https://aclanthology.org/2021.emnlp-main.300/}
}

@inproceedings{chen2022convfinqa,
  title = {{C}onv{F}in{QA}: Exploring the Chain of Numerical Reasoning in Conversational Finance Question Answering},
  author = {Chen, Zhiyu and Li, Shiyang and Smiley, Charese and Ma, Zhiqiang and Shah, Sameena and Wang, William Yang},
  booktitle = {Proceedings of the 2022 Conference on Empirical Methods in Natural Language Processing},
  pages = {6279--6292},
  year = {2022},
  month = dec,
  address = {Abu Dhabi, United Arab Emirates},
  publisher = {Association for Computational Linguistics},
  doi = {10.18653/v1/2022.emnlp-main.421},
  url = {https://aclanthology.org/2022.emnlp-main.421/}
}

@inproceedings{chen2024humans,
  title = {Humans or {LLM}s as the Judge? A Study on Judgement Bias},
  author = {Chen, Guiming Hardy and Chen, Shunian and Liu, Ziche and Jiang, Feng and Wang, Benyou},
  booktitle = {Proceedings of the 2024 Conference on Empirical Methods in Natural Language Processing},
  pages = {8301--8327},
  year = {2024},
  month = nov,
  address = {Miami, Florida, USA},
  publisher = {Association for Computational Linguistics},
  doi = {10.18653/v1/2024.emnlp-main.474},
  url = {https://aclanthology.org/2024.emnlp-main.474/}
}

@inproceedings{es2024ragas,
  title = {{RAGA}s: Automated Evaluation of Retrieval Augmented Generation},
  author = {Es, Shahul and James, Jithin and Espinosa Anke, Luis and Schockaert, Steven},
  booktitle = {Proceedings of the 18th Conference of the European Chapter of the Association for Computational Linguistics: System Demonstrations},
  pages = {150--158},
  year = {2024},
  month = mar,
  address = {St. Julians, Malta},
  publisher = {Association for Computational Linguistics},
  doi = {10.18653/v1/2024.eacl-demo.16},
  url = {https://aclanthology.org/2024.eacl-demo.16/}
}

@inproceedings{liu2024agentbench,
  title = {{A}gent{B}ench: Evaluating {LLM}s as Agents},
  author = {Liu, Xiao and Yu, Hao and Zhang, Hanchen and Xu, Yifan and Lei, Xuanyu and Lai, Hanyu and Gu, Yu and Ding, Hangliang and Men, Kaiwen and Yang, Kejuan and Zhang, Shudan and Deng, Xiang and Zeng, Aohan and Du, Zhengxiao and Zhang, Chenhui and Shen, Sheng and Zhang, Tianjun and Su, Yu and Sun, Huan and Huang, Minlie and Dong, Yuxiao and Tang, Jie},
  booktitle = {The Twelfth International Conference on Learning Representations},
  year = {2024},
  url = {https://openreview.net/forum?id=zAdUB0aCTQ}
}

@inproceedings{liu2023geval,
  title = {{G}-Eval: {NLG} Evaluation using {GPT}-4 with Better Human Alignment},
  author = {Liu, Yang and Iter, Dan and Xu, Yichong and Wang, Shuohang and Xu, Ruochen and Zhu, Chenguang},
  booktitle = {Proceedings of the 2023 Conference on Empirical Methods in Natural Language Processing},
  pages = {2511--2522},
  year = {2023},
  month = dec,
  address = {Singapore},
  publisher = {Association for Computational Linguistics},
  doi = {10.18653/v1/2023.emnlp-main.153},
  url = {https://aclanthology.org/2023.emnlp-main.153/}
}

@inproceedings{ou2025agentdiagnose,
  title = {{A}gent{D}iagnose: An Open Toolkit for Diagnosing {LLM} Agent Trajectories},
  author = {Ou, Tianyue and Guo, Wanyao and Gandhi, Apurva and Neubig, Graham and Yue, Xiang},
  booktitle = {Proceedings of the 2025 Conference on Empirical Methods in Natural Language Processing: System Demonstrations},
  pages = {207--215},
  year = {2025},
  month = nov,
  address = {Suzhou, China},
  publisher = {Association for Computational Linguistics},
  doi = {10.18653/v1/2025.emnlp-demos.15},
  url = {https://aclanthology.org/2025.emnlp-demos.15/}
}

@article{verga2024replacing,
  title = {Replacing Judges with Juries: Evaluating {LLM} Generations with a Panel of Diverse Models},
  author = {Verga, Pat and Hofstatter, Sebastian and Althammer, Sophia and Su, Yixuan and Piktus, Aleksandra and Arkhangorodsky, Arkady and Xu, Minjie and White, Naomi and Lewis, Patrick},
  journal = {arXiv preprint arXiv:2404.18796},
  year = {2024},
  doi = {10.48550/arXiv.2404.18796},
  url = {https://arxiv.org/abs/2404.18796}
}

@inproceedings{xie2024finben,
  title = {{F}in{B}en: A Holistic Financial Benchmark for Large Language Models},
  author = {Xie, Qianqian and Han, Weiguang and Chen, Zhengyu and Xiang, Ruoyu and Zhang, Xiao and He, Yueru and Xiao, Mengxi and Li, Dong and Dai, Yongfu and Feng, Duanyu and Xu, Yijing and Kang, Haoqiang and Kuang, Ziyan and Yuan, Chenhan and Yang, Kailai and Luo, Zheheng and Zhang, Tianlin and Liu, Zhiwei and Xiong, Guojun and Deng, Zhiyang and Jiang, Yuechen and Yao, Zhiyuan and Li, Haohang and Yu, Yangyang and Hu, Gang and Huang, Jiajia and Liu, Xiao-Yang and Lopez-Lira, Alejandro and Wang, Benyou and Lai, Yanzhao and Wang, Hao and Peng, Min and Ananiadou, Sophia and Huang, Jimin},
  booktitle = {Advances in Neural Information Processing Systems},
  volume = {37},
  year = {2024},
  url = {https://proceedings.neurips.cc/paper_files/paper/2024/hash/adb1d9fa8be4576d28703b396b82ba1b-Abstract-Datasets_and_Benchmarks_Track.html},
  doi = {10.52202/079017-3033}
}

@inproceedings{zheng2023judging,
  title = {Judging {LLM}-as-a-Judge with {MT}-Bench and Chatbot Arena},
  author = {Zheng, Lianmin and Chiang, Wei-Lin and Sheng, Ying and Zhuang, Siyuan and Wu, Zhanghao and Zhuang, Yonghao and Lin, Zi and Li, Zhuohan and Li, Dacheng and Xing, Eric P. and Zhang, Hao and Gonzalez, Joseph E. and Stoica, Ion},
  booktitle = {Advances in Neural Information Processing Systems},
  volume = {36},
  year = {2023},
  url = {https://proceedings.neurips.cc/paper_files/paper/2023/hash/91f18a1287b398d378ef22505bf41832-Abstract-Datasets_and_Benchmarks.html}
}

@article{chen2025finfm,
  title={Advancing Financial Engineering with Foundation Models: Progress, Applications, and Challenges},
  author={Chen, Liyuan and Liu, Shuoling and Yan, Jiangpeng and Wang, Xiaoyu and Liu, Henglin and Li, Chuang and Jiao, Kecheng and Ying, Jixuan and Liu, Yang Veronica and Yang, Qiang and Li, Xiu},
  journal={arXiv preprint arXiv:2507.18577},
  year={2025}
}

@article{liu2025deepseek,
  title={When {DeepSeek-R1} meets financial applications: benchmarking, opportunities, and limitations},
  author={Liu, Shuoling and Chen, Liyuan and Yan, Jiangpeng and Jiang, Yuhang and Wang, Xiaoyu and Li, Xiu and Yang, Qiang},
  journal={Frontiers of Information Technology \& Electronic Engineering},
  volume={26},
  number={10},
  pages={1862--1870},
  year={2025},
  doi={10.1631/FITEE.2500227}
}

\end{document}